\documentclass[conference]{IEEEtran}
\IEEEoverridecommandlockouts
\usepackage{cite}
\usepackage{amsmath,amssymb,amsfonts}
\usepackage{algorithmic}
\usepackage{graphicx}
\usepackage[backref]{hyperref} 
\usepackage{textcomp}
\usepackage{xcolor}
\usepackage{booktabs}
\usepackage{algorithm}
\usepackage{algorithmic}
\usepackage{amsmath}
\usepackage{bm}
\def\BibTeX{{\rm B\kern-.05em{\sc i\kern-.025em b}\kern-.08em
    T\kern-.1667em\lower.7ex\hbox{E}\kern-.125emX}}
\begin{document}

\title{VGAT: A Cancer Survival Analysis Framework Transitioning from Generative Visual Question Answering to Genomic Reconstruction  \\

\author{
    \IEEEauthorblockN{Zizhi Chen$^{1,2,\dagger}$, Minghao Han$^{1,2,\dagger}$, Xukun Zhang$^{1,2}$, Shuwei Ma$^{1,2}$, Tao Liu$^{1,2}$, Xing Wei$^{1,2}$, Lihua Zhang$^{1,2,3,4,*}$}
    \thanks{$^\dagger$Equal contribution. $^*$Corresponding author. }
    \thanks{This project was funded by the National Natural Science Foundation of China 82090052.}
    \IEEEauthorblockA{$^1$ Academy for Engineering and Technology, Fudan University, Shanghai, China}
    \IEEEauthorblockA{$^2$ Institute of Metaverse \& Intelligent Medicine, Fudan University, Shanghai, China}
    \IEEEauthorblockA{$^3$ Engineering Research Center of AI and Robotics, Ministry of Education, Shanghai, China}
    \IEEEauthorblockA{$^4$ Jilin Provincial Key Laboratory of Intelligence Science and Engineering, Changchun, China}
    \IEEEauthorblockA{\texttt{chenzz24@m.fudan.edu.cn, mhhan22@m.fudan.edu.cn, lihuazhang@fudan.edu.cn}}
}
}

\maketitle

\begin{abstract}
Multimodal learning combining pathology images and genomic sequences enhances cancer survival analysis but faces clinical implementation barriers due to limited access to genomic sequencing in under-resourced regions. To enable survival prediction using only whole-slide images (WSI), we propose the Visual-Genomic Answering-Guided Transformer (VGAT), a framework integrating Visual Question Answering (VQA) techniques for genomic modality reconstruction. By adapting VQA's text feature extraction approach, we derive stable genomic representations that circumvent dimensionality challenges in raw genomic data. Simultaneously, a cluster-based visual prompt module selectively enhances discriminative WSI patches, addressing noise from unfiltered image regions. Evaluated across five TCGA datasets, VGAT outperforms existing WSI-only methods, demonstrating the viability of genomic-informed inference without sequencing. This approach bridges multimodal research and clinical feasibility in resource-constrained settings. The code link is \url{https://github.com/CZZZZZZZZZZZZZZZZZ/VGAT}.

\end{abstract}


\section{Introduction}
Cancer survival analysis~\cite{cox}~\cite{survival} is a complex ordinal regression task aimed at estimating the relative risk of mortality for cancer prognosis. In recent years, medical image analysis has made significant progress, driven by the success of deep learning techniques. Consequently, an increasing number of researchers are working to establish the connection between imaging features and survival outcomes.

To improve analysis accuracy, the pathology examination will be conducted to sample lesion tissues and acquire pathology images, also known as whole slide images (WSIs), which providing information about microscopic changes in tumor cells and their microenvironment. To identify tumor-related regions within a large number of WSIs, Zhu \textit{et al.}~\cite{MIL_survial} proposed multi-instance learning (MIL) method, which has shown significant effectiveness and has become a paradigm in survival analysis~\cite{atmil,AttnMISL,pmil,transmil,dtfd,dsmil,rrt,PANTHER}. Recently, with the rapid advancement of high-throughput sequencing technologies, large-scale genomic datasets have provided an unprecedented opportunity to understand survival events from a molecular perspective~\cite{SNN}~\cite{bulkrnabert}.

\begin{figure}[tp]
\centerline{\resizebox{3.4in}{1.8in}{\includegraphics{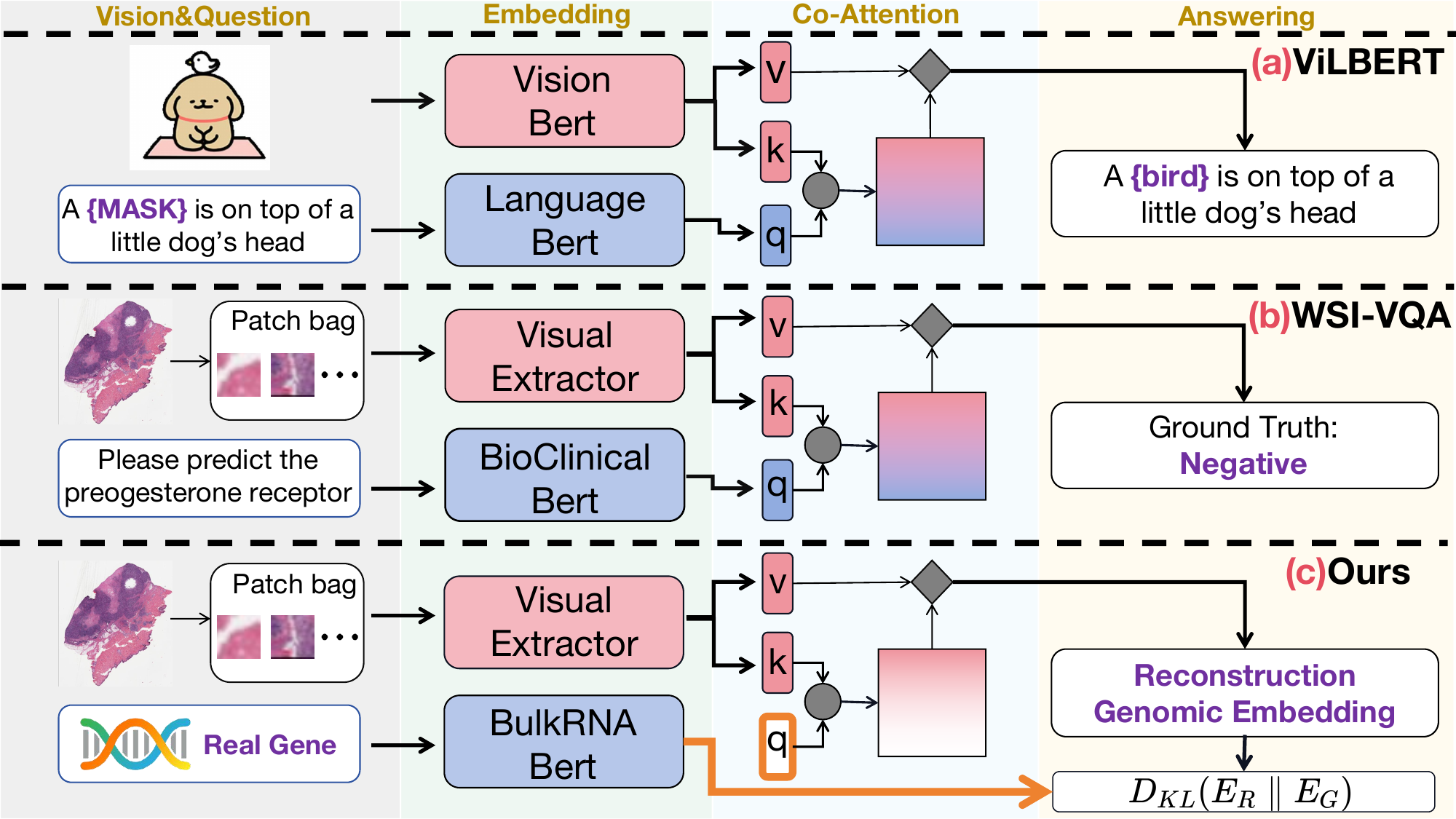}}}
\vspace{-0.5em}
\caption{\textbf{a}: Natural Image Visual Question Answering. \textbf{b}: WSI Visual Question Answering. \textbf{c}: Our Method to Reconstruct Genomics. }
\label{fig1}
\vspace{-1em}
\end{figure}

The gradual abundance of these two types of data resources and technological advancements has also brought the wave of multimodality to survival analysis tasks. Inspired by Visual Question Answering (VQA) tasks~\cite{vse++,SCAN,clip,ViLBERT,vilt}, Chen \textit{et al.}~\cite{MCAT} introduced a co-attention module that primarily integrates pathology and genomic data in multimodal learning. By offering different perspectives for patient stratification and treatment decisions, it not only enhances performance on most cancer datasets but also serves as a cornerstone for subsequent multimodal learning approaches\cite{porpoise,MOTCat,CMTA,mome,PIBD}. However, for the majority of patients, especially those in underdeveloped regions, genomic sequencing remains prohibitively expensive. The disconnect between the demand for multimodal approaches and the current inadequacy of medical resources has impeded the global dissemination and implementation of multimodal survival analysis models.

The strategy of maximizing the utility of existing historical genomic data to reconstruct genomics for complete inference based solely on WSI has been preliminarily validated in studies such as G-HANet and other studies~\cite{G-Hanet}~\cite{MIDLshort24}. In other words, compared to traditional unimodal and multimodal tasks, we aim to introduce a multimodal training and unimodal inference strategy. However, this inevitably involves genomic reconstruction based on historical genomic data. Existing work has recognized the poor performance of reconstructing raw gene sequences and has adopted gene screening methods~\cite{diffseq} to circumvent this issue, but there is still substantial room for improvement. We suggest that understanding the differences between the current reconstruction task and traditional multimodal survival analysis is essential to address this challenge.

To this end, the earliest source of inspiration for the cross-attention mechanism in survival analysis was revisited: VQA tasks. We found that to effectively apply a sufficiently powerful cross-attention mechanism to the patch-level weighting of gigapixel WSI images ($100,000 \times 100,000$ pixels), the processing pipeline for the genomic modality has not followed the framework of VQA models, particularly generative VQA. Instead, it has only utilized co-attention module, which often apply overly simplistic processing (SNN)~\cite{SNN} to the raw genomic sequences, resulting in highly numerical-specific gene features. While this approach has advantages in the fusion of real genes and pathology modalities, the high dimensionality and lack of robustness in genomic reconstruction and subsequent multimodal integration have become significant challenges. Therefore, as illustrated in Fig. \ref{fig1}, we reviewed the overall framework of generative VQA in both natural and pathological image domains~\cite{ViLBERT}~\cite{WSI-VQA}. We found that using transformers, particularly the pre-trained BERT architecture, effectively extracts textual features. However, it is important to note that text and genomic modalities differ, and whether similar methods can be applied to genomic data remains an open question. Nevertheless, studies have shown that pre-training BERT on raw genomic information yields beneficial features for downstream tasks such as survival analysis and tumor classification~\cite{bulkrnabert}. Therefore, we gradually validate the feasibility of a framework that is structurally similar to VQA and tailored for genomic reconstruction.

On the other hand, the fact that cancer cell-related regions in pathology images account for only about $20\%$ of the image content dictates~\cite{PIBD} that the task of genomic reconstruction cannot be performed using overly redundant visual prompts. This issue has already been a challenge in VQA task on pathological images~\cite{WSI-VQA}, and some multi-instance learning works have only employed random or hard cluster assignment downsampling methods~\cite{AttnMISL}~\cite{pmil}, without screening based on the visual features of individual patches and their correlation with the tumor microenvironment. Relevant studies~\cite{PANTHER}~\cite{MMP} have demonstrated that statistical parameters derived from Gaussian Mixture Models (GMM) and Expectation-Maximization (EM)~\cite{EM}~\cite{em2} algorithms can effectively align with the genomic modality through operations such as Optimal Transport (OT)~\cite{OT}~\cite{MOTCat} to achieve multimodal integration. This suggests that fully considering the heterogeneity of patches can better capture the underlying relationship with genomic data. Additionally, a large number of patches unrelated to gene expression used as visual prompts would unnecessarily complicate the visual-genomic question-answering task. Therefore, a novel patch selection mechanism is essential to enhance the effectiveness of this process.

Based on this, we have made a new attempt at multimodal training and unimodal inference and proposed the \textbf{V}ision \textbf{G}enomic \textbf{A}nswering-Guided \textbf{T}ransformer (VGAT). Our contributions are as follows:
(1) We propose a novel framework better suited for genomic reconstruction tasks, replacing raw gene sequences with robust genomic embeddings. Based on this, we have designed a new Visual Genomic Answering (VGA) module.
(2) We first introduce patch selection based on unsupervised algorithms in the context of prognostic tasks, proposing the EM-based Slide Embedding (ESE) module. The EM algorithm calculates the contribution of each patch, and we prove that it can condense sufficiently great visual prompts content.
(3) Extensive experiments were conducted on five public TCGA datasets to evaluate the effectiveness of the proposed model. The experimental results demonstrate that the model consistently outperforms existing methods.

\begin{figure*}[tp]
    \centering
    \includegraphics[width=0.98\linewidth]{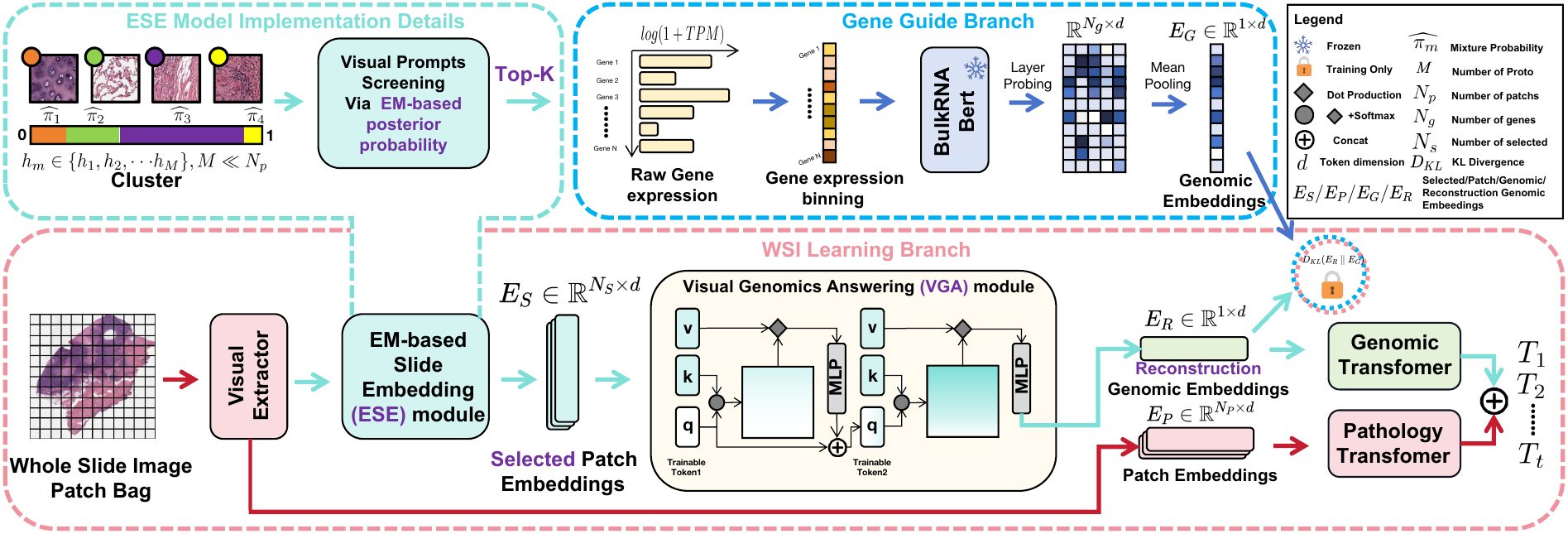}
    \vspace{-0.5em}
    \caption{Overview of \textbf{V}ision \textbf{G}enomic \textbf{A}nswering-Guided \textbf{T}ransformer (VGAT). The pathology images and genomic data are processed into corresponding embeddings. During training, genomic reconstruction is performed using selected visual embeedings and genomic embeddings. At inference, only the pathology embeddings are used for genomic reconstruction and survival analysis.}
    \label{fig2}
    \vspace{-1em}
\end{figure*}

\section{Proposed method}

\subsection{Preprocessing for Bag Construction}\label{AA}
\vspace{0.5pt}\noindent\textbf{Visual Embeddings.}
WSIs inherently possess gigapixel resolutions, presenting significant hardware-level challenges for deep learning-based analyses. Building on previous studies, we employ the weakly supervised MIL strategy~\cite{MIL_survial}, through which each WSI can be formulated as multiple instances assigned with one bag-level label. We initially utilize the automated segmentation algorithm in ~\cite{clam} to distinguish the tissue regions over 20× magnification from the background and then crop the informative tissues into a set of  $256\times256 $  patches \(X^P = \{x^p_i\}_{i=1}^{N_p}\). Eventually, the \(N_p\) cropped patches are converted into \(d\)-dimensional feature vectors~\cite{resnet} through a pre-trained encoder \(f_p(\cdot)\) as:
\begin{equation}
\small
E_P = \{f_p(x^p_i) : x^p_i \in X^p\} \in \mathbb{R}^{N_p \times d}.
\end{equation}

\noindent\textbf{Genomic Embeddings.}
In the context of RNA-seq data, each sequence is composed of real values corresponding to gene expression (in Transcript Per Million, or TPM, units). First, we need to pair the Tokens with the real values of genes after linear transformation and then complete the preprocessing and adjustment of feature embedding dimensions according to the requirements of the pre-trained BulkRNABert~\cite{bulkrnabert}. To avoid being restricted to the embedding dimension chosen by Gene2Vec~\cite{gene2vec}, the obtained genomic embeddings is passed through an additional linear layer to reach effective embeddings size \(N_{g}\). Ultimately, as shown in Fig. \ref{fig2}, we take the mean of the output tokens as the genomic embeddings $E_G$.

\subsection{EM-based Slide Embedding module}\label{AA}
The proportion of patches in pathology images that are related to gene expression is extremely limited, and they need to be filtered to prevent genomic reconstruction from focusing on unmeaning visual prompts. We introduce a patch-level screening module and do not utilize all pathology features as visual prompts for the genomic reconstruction task. To this end, we introduce the Gaussian Mixture Model (GMM) and the Expectation-Maximization (EM) algorithm, which allows us to measure the contribution of each component to the data point, enabling the quantification of each patch's importance. Based on this, we establish an EM-based Slide Embedding module for patch screening.

\vspace{0.5pt}\noindent\textbf{Cluster.}
We randomly sampled $p$ patches from TCGA of the same cancer type and performed K-means clustering~\cite{kmeans} to generate $C_h$ different clusters. The centroid vectors of these clusters, denoted as $\left \{ \mathbf{a}_{c} \right \} _{c=1}^{C_h} $, are used for subsequent calculations. To achieve a balance between effectiveness and computational load, $C_h$ is set to 16, and $p$ is set to 1,000,000.

\vspace{1pt}\noindent\textbf{Gaussian Mixture Model.}
The GMM is used to measure the weight of each patch $\mathbf{z}_{i}$; the probability distribution for $\mathbf{z}_{i}$ is:

\begin{equation}
\begin{aligned}
\small
p(\mathbf{z}_{i}; \theta) &= \sum_{c=1}^{C_h} p(c_i=c) \cdot p(\mathbf{z}_{i} | c_i=c) \\
&= \sum_{c=1}^{C_h} \pi_c \cdot \mathcal{N}(\mathbf{z}_{i}; \boldsymbol{\mu}_c, \Sigma_c),
\end{aligned}
\tag{2}
\end{equation}
where $c_i$ denotes the mixture identity of $\mathbf{z}_{i}$ and $\theta = \{\pi_c, \mu_c, \Sigma_c\}_{c=1}^{C_h}$ denotes mixture probability, mean, and diagonal covariance. The posterior distribution $p(c_i = c | \mathbf{z}_{i})$ indirectly represents the distance between $\mathbf{z}_{i}$ and $\mathbf{a}_{c}$.

\vspace{1pt}\noindent\textbf{EM-based Posterior Probability.}
The EM algorithm can iteratively estimate the parameters of a GMM through the expectation step (E-step) and the maximization step (M-step). We utilize the intermediate variables obtained in the E-step for subsequent use, as given by the following formula:
\begin{equation}
\small
\gamma(z_{i,c}) = \frac{\pi_c \cdot \mathcal{N}(z_{i} ; \mu_c, \Sigma_c)}{\sum_{k=1}^{C_h} \pi_k \cdot \mathcal{N}(z_{i} ; \mu_k, \Sigma_k)},\tag{3}
\end{equation}
where $\gamma(z_{i,c})$ denotes the posterior probability of the $i$-th patch being generated by the $c$-th cluster. This step is also referred to as responsibility computation, as it quantifies the contribution of each Gaussian component to the data point. We consider $\gamma(z_{i,c})$ as a reference indicator of the importance and specificity of a patch.

\begin{table*}[tp]
\centering
\caption{Performance comparison with state-of-the-art genome-based/WSI-based inference methods and histo-genomic multi-modal methods on TCGA datasets in terms of c-index. The `Patho.’ (WSI) and `Geno.’ (genomic data) indicate the data utilized during inference. We label the best WSI-based inference model with \textcolor{red}{red} color and \textbf{highlight} the best performance in bold format.}
\vspace{-0.5em}
\resizebox{0.95\textwidth}{!}{
\begin{tabular}{l|cc|ccccccc}
\toprule
\textbf{Methods} & \textbf{Patho.} & \textbf{Geno.}   & \textbf{BLCA}(n=373) & \textbf{BRCA}(n=956) & \textbf{GBMLGG}(n=480) & \textbf{LUAD}(n=569) & \textbf{UCEC} (n=453) & \textbf{Overall} \\
\midrule
SNN~\cite{SNN}  &  &$\checkmark$  &   0.583 $\pm$ 0.021  &  0.609 $\pm$ 0.046 &  0.826 $\pm$ 0.041 & 0.614 $\pm$ 0.039 & 0.684 $\pm$ 0.081 &0.663  \\
SNNTrans~\cite{SNN}~\cite{transmil}  &  & $\checkmark$ &  0.633 $\pm$ 0.043  &0.656 $\pm$ 0.042  & 0.836 $\pm$ 0.038 &0.634 $\pm$ 0.028 &0.713 $\pm$ 0.059 &0.694  \\
BulkRNABert~\cite{bulkrnabert}  &  & $\checkmark$ & 0.631 $\pm$ 0.066 & 0.583 $\pm$ 0.024& 0.866 $\pm$ 0.029 &0.644 $\pm$ 0.028 & \textbf{0.758 $\pm$ 0.101}& 0.696 \\
\midrule
AttnMIL~\cite{atmil}  & $\checkmark$ &  &   0.560 $\pm$ 0.031 & 0.579 $\pm$ 0.074 & 0.737 $\pm$ 0.034 & 0.577 $\pm$ 0.046 & 0.646 $\pm$ 0.089 & 0.620 \\
CLAM-SB~\cite{clam}  & $\checkmark$ &  & 0.623 $\pm$ 0.013  & 0.573 $\pm$ 0.073 & 0.798 $\pm$ 0.028& 0.649 $\pm$ 0.031 & 0.693 $\pm$ 0.058 & 0.667 \\
TransMIL~\cite{transmil}  & $\checkmark$ &  &  0.604 $\pm$ 0.051  &0.614 $\pm$ 0.087&0.766 $\pm$ 0.029 & 0.650 $\pm$ 0.046 & 0.711 $\pm$ 0.081& 0.665 \\
DSMIL~\cite{dsmil}  & $\checkmark$ &  &   0.648 $\pm$ 0.023 & 0.571 $\pm$ 0.042 & 0.800 $\pm$ 0.040 & 0.626 $\pm$ 0.031 & 0.685 $\pm$ 0.046 & 0.666 \\
DTFD-MIL~\cite{dtfd}  & $\checkmark$ &  &   0.607 $\pm$ 0.054 & 0.560 $\pm$ 0.060 & 0.802 $\pm$ 0.029 & 0.620 $\pm$ 0.044 & 0.668 $\pm$ 0.060 & 0.651\\
PANTHER~\cite{PANTHER} & $\checkmark$ &  &   0.586 $\pm$ 0.027  &  0.642 $\pm$ 0.049 &  0.742 $\pm$ 0.026 & 0.631 $\pm$ 0.027 & 0.644 $\pm$ 0.063 &0.649  \\
RRT-MIL~\cite{rrt}  & $\checkmark$ &  &   0.634 $\pm$ 0.025 & 0.647 $\pm$ 0.050 & 0.806 $\pm$ 0.029 & 0.630 $\pm$ 0.039 & 0.697 $\pm$ 0.052 & 0.683 \\
\midrule
MCAT~\cite{MCAT}  & $\checkmark$ & $\checkmark$ & 0.663 $\pm$ 0.061 &  0.655 $\pm$ 0.022 &  0.852 $\pm$ 0.037 &  0.682 $\pm$ 0.033 &  0.698 $\pm$ 0.071 & 0.710 \\
CMTA~\cite{CMTA}  & $\checkmark$ & $\checkmark$ & 0.679 $\pm$ 0.042& 0.659 $\pm$ 0.029 & 0.848 $\pm$ 0.031 & \textbf{0.687 $\pm$ 0.027}& 0.721 $\pm$ 0.098 & 0.719 \\
MOTCat~\cite{MOTCat}  & $\checkmark$ & $\checkmark$ & \textbf{0.685 $\pm$ 0.053}  & \textbf{0.675 $\pm$ 0.064} &0.848 $\pm$ 0.045  & 0.680 $\pm$ 0.026 & 0.714 $\pm$ 0.066 & \textbf{0.720} \\
MOME~\cite{mome} &  $\checkmark$&$\checkmark$  &  0.668 $\pm$ 0.056  &0.658 $\pm$ 0.047  & 0.853 $\pm$ 0.038 & 0.670 $\pm$ 0.018 & 0.714 $\pm$ 0.074& 0.713 \\
MultiBert~\cite{bulkrnabert}~\cite{transmil} &  $\checkmark$&$\checkmark$  &  0.654 $\pm$ 0.062  &0.643 $\pm$ 0.053  & \textbf{0.872 $\pm$ 0.032} & 0.655 $\pm$ 0.046 & 0.742 $\pm$ 0.062& 0.713 \\
\midrule
G-HANet~\cite{G-Hanet} &  $\checkmark$&\textbf{Training Only}  &  0.648 $\pm$ 0.019  & 0.633 $\pm$ 0.037 & 0.810 $\pm$ 0.034 & 0.632 $\pm$ 0.028&\textcolor{red}{0.737 $\pm$ 0.035} & 0.692 \\
VGAT(Ours)&  $\checkmark$&\textbf{Training Only}  & \textcolor{red}{0.671 $\pm$ 0.027}   &\textcolor{red}{0.658 $\pm$ 0.081}  & \textcolor{red}{0.812 $\pm$ 0.028} &\textcolor{red}{0.655 $\pm$ 0.050} & 0.721 $\pm$ 0.057& \textcolor{red}{0.703} \\
\bottomrule
\end{tabular}
}
\vspace{-0.5em}
\label{table1}
\end{table*}

\begin{figure*}[tp]
    \centering
    \includegraphics[width=0.93\linewidth]{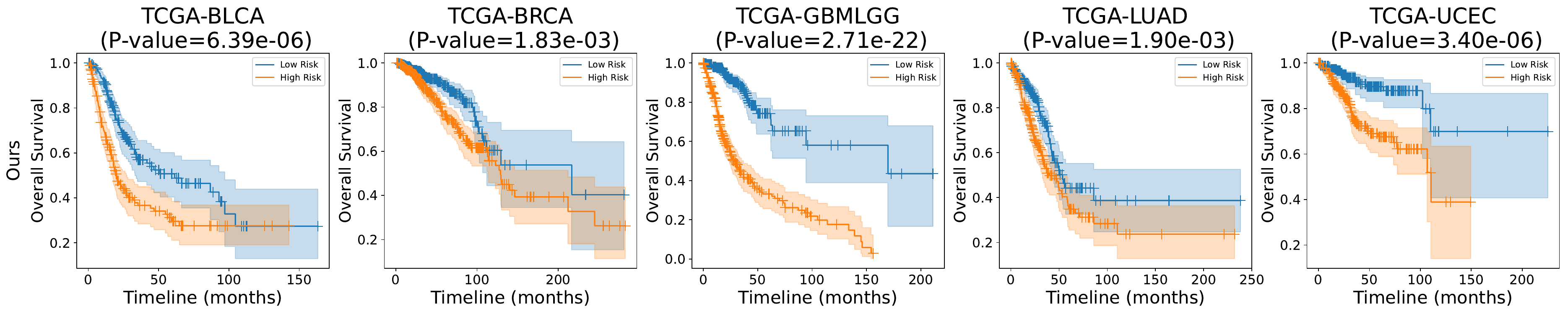}
    \vspace{-1em}
    \caption{According to predicted risk scores, all patients are stratified into low risk group (blue) and high risk group (orange). Then, we utilize Kaplan-Meier analysis and Logrank test (p-value) to measure the statistical significance between low risk group and high risk group.}
    \label{fig3}
    \vspace{-1em}
\end{figure*}

\vspace{1pt}\noindent\textbf{Visual Prompts Screening.}
We consider the class corresponding to the maximum value of the posterior probability $\gamma(z_{i,c})$, denoted as $\gamma(z_i)$, as the classification result:

\begin{equation}
\small
\gamma(z_{i}) = \max_{c \in \{1, 2, \dots, C_h\}} \gamma(z_{i,c}).\tag{4}
\end{equation}

A unique threshold gating module was designed to perform multi-scale TOP-K-based patch selection, which is based on the number of patches at the cluster scale and the category posterior probabilities at the patch scale. This module selects patches that are representative and distinctive to form the visual prompts for genomic reconstruction. Please refer to the \textbf{Supplementary Materials} for the module implementation.

\subsection{Visual Genomic Answering module  }\label{AA}
In our framework, we refer to the visual prompts embeddings processed by the ESE module as $E_S$ and the genomic embeddings as $E_G$. To achieve the purpose of genomic reconstruction, we introduce learnable tokens called $E_L$. Similar to most parallel VQA frameworks, we use a cross-modal attention mechanism to complete the fine-grained interaction between the learnable tokens and visual embeddings. As shown in the following formula:

\begin{equation}
\begin{split}
\text{CoAtt}(E_S,E_L) 
&= \text{softmax}\left(\frac{W_Q E_L E_S^T W_K^T}{\sqrt{d}}\right)W_V E_S,\\
&= \textbf{A}W_V E_S,
\end{split}
\tag{5}
\end{equation}
where $W_Q$,$W_K$,$W_V$ $\in$ $\mathbb{R}^{d \times d}$ are trainable weight matrices multiplied by the visual embeddings and learnable tokens. The co-attention module \textbf{A} $\in$ $\mathbb{R}^{N_{s} \times N_{L}}$ uncovers the fine-grained patch and learnable token similarity, revealing the underlying mapping between gene data and visual prompts. Subsequently, following the cross-modal attention, a GELU non-linearity and a residual-architecture MLP layer are employed. We refer to the final reconstructed genomic embeddings as $E_R$.

Ultimately, we will constrain the genomic reconstructed based on visual prompts using the Kullback-Leibler (KL) divergence~\cite{KL} against the real genomic embeddings that have already undergone preprocessing. This will allow the learnable tokens to gradually learn the latent connections between visual prompts and genes, akin to “describing images” but for “recognizing genes from images”. The formula for KL divergence is as follows:
\begin{equation}
\small
\mathcal{L}_{KL} = \mathcal{KL}(E_R \parallel E_G) = \sum_{x\in d} r(x) \log\left(\frac{r(x)}{g(x)}\right),\tag{6}
\end{equation}
$r(x)$, $g(x)$ are the probability density functions of $E_R$, $E_S$.

\subsection{Multimodal Survival Analysis from Gene Reconstruction}\label{AA}

\vspace{1pt}\noindent\textbf{Pathology and Genomic Transformer.} Whether dealing with pathology image embeddings $E_P$ or reconstructed genomic embeddings $E_G$, we can summarize it as $E = \left \{E_1, E_2, \cdot \cdot \cdot , E_{N}  \right \} $ and by adding a Token $E_0$ as a Classification (CLS) to aggregate all feature information, we refer to it as $E^{(0)} = \left \{E_0^{(0)}, E_1^{(0)}, \cdot \cdot \cdot , E_{N}^{(0)}   \right \} \in \mathbb{R}^{(N+1)\times d  } $ as the input for the transformer encoder. We apply two
self-attention layers to perform information integration. Additionally, in the pathology encoder, there is another PPEG (Pyramid Position Encoding Generator)~\cite{transmil} module to explore the correlations among different patches. The computation of the pathology encoder can be formulated as follows:
\begin{equation}
\small
E^{(1)} = \text{SelfAtt}(\text{LN}(E^{(0)}))+E^{(0)},\tag{7}
\end{equation}
\begin{equation}
\small
E^{(2)} = \text{PPEG}(E^{(1)}),\tag{8} \label{equ9}
\end{equation}
\begin{equation}
\small
E^{(3)} = \text{SelfAtt}(\text{LN}(E^{(2)}))+E^{(2)},\tag{9}
\end{equation}
where SelfAtt denotes Self-attention, and LN denotes Layer Norm. The Genomic Encoder does not involve the process described in equation \ref{equ9}.

\vspace{1pt}\noindent\textbf{Survival Prediction.}
We concat CLS of $E_P^{(3)}$ and $E_R^{(2)}$ to an MLP to derive the final risk scores. This simplified multimodal approach is deliberately chosen to strongly link results with genomic reconstruction quality, highlighting the VGA framework's potential. The risk scores are then subjected to supervision by applying the NLL-loss~\cite{survival}. This constraint is formulated in conjunction with the previous KL divergence during training, and further details regarding survival analysis can be found in the \textbf{Supplementary Materials}.

\section{Experiments}
\subsection{Implementation Details}\label{AA}
VGAT is implemented in PyTorch and trained on a workstation with 4 NVIDIA A6000 GPUs. The Adam optimizer was used with a learning rate of 2e-4 (1e-5 for the BRCA), a weight decay of 1e-5, and to ensure fairness and account for the complexity of genomic reconstruction learning, all models were trained for 100 epochs. To fairly evaluate different methods, identical loss functions, seeds, and ResNet-50~\cite{resnet} feature extractors were employed across all approaches. 

\subsection{Datasets and Evaluation Metrics}\label{AA}

To validate our proposed method, we utilized the five most enormous cancer datasets from The Cancer Genome Atlas (TCGA). This public consortium provides comprehensive diagnostic whole-slide images (WSIs) and genomic data, including labelled survival times and censorship statuses. Specifically, cancer types and data quantities are detailed in Table \ref{table1}. The predictive performance of our method in correctly ranking the patient risk scores relative to overall survival was evaluated using the 5-fold cross-validated concordance index (C-Index) for each cancer dataset.

\subsection{Comparisons with Other Methods}\label{AA}
We compare our method against the unimodal baselines and the multimodal SOTA methods as follows:

\vspace{1pt}\noindent\textbf{Unimodal Baseline.}
For genomic data, we utilized both \textbf{SNN}~\cite{SNN} and \textbf{SNNTrans}~\cite{SNN}~\cite{transmil}, which have been commonly employed in previous studies for predicting survival outcomes in TCGA. Subsequently, we adopted \textbf{BlukRNABert}~\cite{bulkrnabert}, a novel approach that extracts robust genomic embeddings, providing a more suitable and advanced method for survival analysis. For histology, we campare the SOTA MIL methods \textbf{AttnMIL}~\cite{atmil}, \textbf{CLAM-SB}~\cite{clam}, \textbf{TransMIL}~\cite{transmil}, \textbf{DSMIL}~\cite{dsmil}, \textbf{DTFD-MIL}~\cite{dtfd}, \textbf{PANTHER}~\cite{PANTHER} and \textbf{RRT-MIL}~\cite{rrt}. The multi-modal training and uni-modal inference work \textbf{G-HANet}~\cite{G-Hanet} has also been compared.

\vspace{1pt}\noindent\textbf{Multimodal SOTA.}
We compare four SOTA methods for multimodal survival analysis including\textbf{ MCAT}~\cite{MCAT}, \textbf{CMTA}~\cite{CMTA}, \textbf{MOTCat}~\cite{MOTCat}, \textbf{MOME}~\cite{mome} and \textbf{MultiBert}~\cite{bulkrnabert}~\cite{transmil}, which is a multimodal approach to replace our $E_S$ with $E_G$.

The results are shown in Table \ref{table1}; it is essential to emphasize that since we did not use the genetic modality in our inference process, a direct comparison with the numerical results of multimodal inference methods is unfair. Experimental results involving all methods are simply presented to demonstrate the potential of our approach.

\begin{table}[tp]
\centering
\caption{comparision of bulkrnabert Embedding with gene sequence}
\vspace{-0.9em}
\resizebox{0.48\textwidth}{!}{%
\begin{tabular}{l|ccc}
\toprule
\textbf{Geno.} &  \textbf{BLCA} & \textbf{BRCA} & \textbf{GBMLGG}\\
\midrule
Original sequence  & 0.613 $\pm$ 0.021  & 0.617 $\pm$ 0.022 &0.797 $\pm$ 0.033 \\
Differential analyzed sequence  & 0.647 $\pm$ 0.028  & 0.617 $\pm$ 0.033 &0.806 $\pm$ 0.023  \\
BulkRNABert Embedding(Ours) & \textbf{0.671 $\pm$ 0.027}  & \textbf{0.658 $\pm$ 0.081} &\textbf{0.812 $\pm$ 0.028}  \\
\bottomrule
\end{tabular}
}
\vspace{-1em}
\label{table2}
\end{table}

\begin{table}[tp]
\centering
\caption{Ablation study assessing the impact of modules in WSI branch. }
\vspace{-0.9em}
\resizebox{0.46\textwidth}{!}{%
\begin{tabular}{cc|ccc}
\toprule
\textbf{ESE} & \textbf{VGA}& \textbf{BLCA} & \textbf{BRCA} & \textbf{GBMLGG}\\
\midrule
 -&-& 0.594 $\pm$ 0.021  & 0.618 $\pm$ 0.041 &0.784 $\pm$ 0.021  \\
-&$\checkmark$ & 0.670 $\pm$ 0.028  & 0.629 $\pm$ 0.063 &0.806 $\pm$ 0.023 \\
$\checkmark$&$\checkmark$ & \textbf{0.671 $\pm$ 0.027}  & \textbf{0.658 $\pm$ 0.081} &\textbf{0.812 $\pm$ 0.028} \\
\bottomrule
\end{tabular}
}
\vspace{-2em}
\label{table3}
\end{table}

\vspace{1pt}\noindent\textbf{VGAT \emph{vs.} Genomic-based Methods.}
Compared to genome-based methods, VGAT demonstrates comprehensive superiority across the BLCA, BRCA, and LUAD datasets, indicating that the visual phenotypic information provided by pathology images for different cancer grades is highly significant in characterizing the tumor microenvironment in these three datasets. On the other hand, the outstanding performance of BulkRNABert on the UCEC and GBMLGG datasets also provides a strong justification for abandoning the original genomic sequence structure.

\vspace{1pt}\noindent\textbf{VGAT \emph{vs.} WSI-Based Inference Method.}
This comparison is divided into two aspects. Firstly, it is compared with the WSI-Based Method. VGAT shows overwhelming advantages on all five datasets, with peak performance improved by $0.7\%$ to $3.5\%$ respectively, improving the mean C-index by $2.9\%$. This demonstrates that the method of maximizing historical genomic data can effectively reconstruct the corresponding genetic information of pathology images, yielding results far superior to previous methods. This is evidenced by the leading performance of VGAT and G-HANet in terms of mean C-index. And the most critical aspect is the lateral comparison test with the same strategy methods: under the strategy of multi-modal training and single-modal inference, VGAT refreshed the best results in four out of five datasets, improving the mean C-index by $1.5\%$.

\begin{table}[tp]
\centering
\caption{comparision of EM-based slide embedding with other methods }
\vspace{-0.9em}
\resizebox{0.46\textwidth}{!}{%
\begin{tabular}{l|ccc}
\toprule
\textbf{slide embedding.} &  \textbf{BLCA} & \textbf{BRCA} & \textbf{GBMLGG}\\
\midrule
-  & 0.670 $\pm$ 0.028  & 0.629 $\pm$ 0.063 &0.806 $\pm$ 0.023  \\
Rand  & 0.657 $\pm$ 0.032  & 0.615 $\pm$ 0.043 &0.809 $\pm$ 0.027  \\
Cluster-based  & 0.670 $\pm$ 0.038  & 0.636 $\pm$ 0.034 &0.812 $\pm$ 0.029  \\
EM-based(Ours)  & \textbf{0.671 $\pm$ 0.027}  & \textbf{0.658 $\pm$ 0.081} &\textbf{0.812 $\pm$ 0.028}  \\
\bottomrule
\end{tabular}
}
\vspace{-2em}
\label{table4}
\end{table}

\vspace{1pt}\noindent\textbf{VGAT \emph{vs.} Multimodal Methods.}
In comparing with multi-modal methods, we first obtained the expected results, with the overall experimental results being only slightly lower than the latest multi-modal approaches (The mean C-index is only slightly less than 1$\%$ lower than the technically mature MCAT). Even more surprising is that VGAT outperformed most multi-modal methods on the UCEC datasets. This suggests the new framework may be a more implicit multi - modal fusion method, with potential for further research to achieve SOTA across all strategies. 

\vspace{1pt}\noindent\textbf{Embedding \emph{vs.} Sequence.}
As shown in Table \ref{table2}, we compared the effects of using $E_G$ as a constraint for genomic reconstruction tasks with the traditional multimodal approach using gene sequences. Based on the unchanged model framework, compared to the original sequence and the differential analysis sequence, the use of $E_G$ improved the C-index mean by $5.2\%$ and $2.4\%$, respectively. (complete experimental results can be found in the \textbf{Supplementary Materials}). This clearly demonstrates the adaptability of the Visual Genomic-Answering to survival analysis, and a more robust representation extraction process enhances the reliability of genomic reconstruction.

\subsection{Ablation Studies}\label{AA} Tables \ref{table3} and \ref{table4} present the results of an ablation study conducted on three datasets to assess the influence of the ESE module and the VGA module. When the VGA module is absent, VGAT degenerates into a WSI-based method similar to TransMIL. When the ESE module is absent, no screening occurs, and $N_S=N_P$. Table \ref{table4} provides more details on the visual embeddings screening strategies. To ensure a fair comparison, the total number of $N_S$ is uniformly set to be the same under different screening strategies. The experimental results show that any module of VGAT is indispensable; reducing or replacing any module will decrease the effectiveness of the experiments. In addition, we compared the effects of different loss function transformations with $E_G$ completion constraints and ultimately concluded that $\mathcal{L}_{KL}$ performed best across four datasets. The complete ablation study results for the five datasets and the comparison histogram of the loss functions are provided in the \textbf{supplementary materials}.

\subsection{Patient Stratification}
\label{AA}
To further validate the effectiveness of VGAT in survival analysis, we divided all patients into low-risk and high-risk groups based on the median of the predicted risk scores from VGAT. Subsequently, we visualized the survival events of all patients using Kaplan-Meier analysis, and the results are presented in Fig. \ref{fig3}. Additionally, we used the Logrank test (p-value) to measure the statistical significance between the low-risk group (blue) and the high-risk group (orange). A p-value of less than or equal to 0.05 is typically considered statistically significant. As shown in the figure, the p-values for all datasets are clearly less than 0.05. For detailed visual comparisons with other
methods, please refer to the \textbf{Supplementary Materials}.

\section{Conclusion} Compared to other methods, the proposed framework can reconstruct genetic information, address the issue of scarce gene data, and achieve better performance in survival prediction and patient stratification tasks.

\section*{Acknowledgment}
This project was funded by the National Natural Science
Foundation of China 82090052.

{\small
\bibliographystyle{IEEEtran}
\bibliography{reference.bib}
}

\section{Appendix}

\subsection{Multi-Scale TOP-K-based Patchs Selected Module}
In the main text, we mentioned the Multi-Scale TOP-K-based Patch Selection Module, which measures the characteristics of patches from two aspects: cluster rarity and posterior probability rarity. This module selects the most representative and distinctive patches as visual prompts. The following is the algorithm \ref{algorithm1} for the module, with the specific setting being that we equally choose patches with the highest and lowest posterior probabilities. We believe that patches closest to the cluster centers and those at the farthest edges of the clusters play significant roles:

\begin{algorithm}[H]
\caption{The TOP-K-based Patchs Selected Algorithm}
\begin{algorithmic}[1]
\REQUIRE $ $\\
$E_P \in \mathbb{R}^{N_p \times d.}:$The pathology images embeddings
\\$N_S\in\mathbb{Z}: \text{The total number of visual prompts set.}$
\ENSURE $ $
\\$E_S \in \mathbb{R}^{N_s \times d.}$:The EM-based Slide Embeddings 

\STATE $probs \gets \text{EM}(E_P)$
\STATE //Calculate the posterior probability between each patch and each cluster and store it in an array.
\STATE $max\_probs, patch\_classes \gets \text{max}(probs)$
\STATE $patch\_map \gets (\text{patch\_classes, max\_probs)}$
\STATE $counts \gets \text{count}(patch\_classes)$
\STATE //Obtain the number of patches in each cluster.
\FOR{$category, count$ \textbf{in} $\text{counts}$}
    \IF{$count < N_S/32$}
        \FOR{$(i, c, p)$ \textbf{in} $patch\_map$}
            \IF{$c == category$}
                \STATE $selected\_indices.append(i)$
                \STATE //If patch classfication are rare, then select all.
            \ENDIF
        \ENDFOR
    \ELSIF{$count \geq N_S/16$}
         \FOR{$(i, c, p)$ \textbf{in} $patch\_map$}
            \IF{$c == category\&\&(p = topK(p_{max}||p_{min} ))$}
                \STATE $selected\_indices.append(i)$
                \STATE //If prob are rare, then select it.
                \ENDIF
        \ENDFOR
    \ENDIF
\ENDFOR
\IF{$\text{len}(selected\_indices) < N_S$}
    \STATE non-overlapping random padding
\ENDIF
\STATE $E_S \gets E_P[selected\_indices]$
\end{algorithmic}
\label{algorithm1}
\end{algorithm}

\subsection{Survival Analysis}
\vspace{1pt}\noindent\textbf{Preliminaries.}
Cancer survival analysis is a complex ordinal regression task aimed at estimating the relative risk of mortality for cancer prognosis. The analysis is completed based on the four data points of the patient.

Let $\mathbb{X}=\left \{ X_1,X_2,\cdot \cdot \cdot X_N \right \}$  represent the quadruple data for $N$ patients. Specifically, the data for any patient $X_i$ can be represented in the format of quadruple: $X_i=\left \{ P_i,G_i,c_i,t_i \right \}$. $P_i$ and $G_i$ represent pathology images and Genomic Data. $t \in \mathbb{R} ^+$ signifies the overall survival time, $c_i \in \{0,1\}$ denotes whether the outcome event is right-censored $(c = 1)$ or not $(c = 0)$. In survival analysis, let $T$ be a continuous random variable for overall survival time. In the training phase, our model integrates the $P_i$ and $G_i$ of any patient, and in the inference phase, it uses only the $P_i$ to estimate the hazard function $h(T = t)$,

\begin{align}
\small
h(T = t)  = \lim_{\partial t \to 0}\frac{P(t\le T \le t+\partial t|T \ge t)}{\partial t}, 
\tag{10}
\end{align}
which can be used to estimate $S_n(t|D_n)$ by integrating over of $h$. The most common method for estimating the hazard function is the Cox Proportional Hazards (CoxPH) model, in which $h$ is parameterized as an exponential linear function,
\begin{align}
\small
h(t|D) & = h_0(t)e^{\theta D},
\tag{11}
\end{align}
 where $h_0$ is the baseline hazard function and $\theta$ are model parameters that describe how the hazard varies with data $D$.  Using deep learning, $\theta$ is the last hidden layer in a neural network and can be optimized using Stochastic Gradient Descent with the Cox partial log-likelihood.

 \normalsize
\begin{table*}[tp]
\centering
\vspace{-1.5em}
\caption{comparision of bulkrnabert Embedding with gene sequence in all datasets}
\resizebox{0.93\textwidth}{!}{%
\begin{tabular}{l|ccccc}
\toprule
\textbf{Geno.} &  \textbf{BLCA} & \textbf{BRCA} & \textbf{GBMLGG}& \textbf{LUAD}& \textbf{UCEC}\\
\midrule
Original sequence  & 0.613 $\pm$ 0.021  & 0.617 $\pm$ 0.022 &0.797 $\pm$ 0.033&0.630 $\pm$ 0.031 & 0.685 $\pm$ 0.060  \\
Differential analyzed sequence  & 0.647 $\pm$ 0.028  & 0.617 $\pm$ 0.033 &0.806 $\pm$ 0.023&0.643 $\pm$ 0.027 & 0.721 $\pm$ 0.034 \\
BulkRNABert Embedding(Ours)  & \textbf{0.671 $\pm$ 0.027}  & \textbf{0.658 $\pm$ 0.081} &\textbf{0.812 $\pm$ 0.028}&\textbf{0.655 $\pm$ 0.050} & \textbf{0.721 $\pm$ 0.057} \\
\bottomrule
\end{tabular}
}
\vspace{-1.5em}
\label{table5}
\end{table*}

\begin{table*}[tp]
\centering
\caption{Ablation study assessing the impact of modules in WSI branch in all datasets. }
\resizebox{0.93\textwidth}{!}{%
\begin{tabular}{cc|ccccc}
\toprule
\textbf{ESE} & \textbf{VGA}& \textbf{BLCA} & \textbf{BRCA} & \textbf{GBMLGG}& \textbf{LUAD}& \textbf{UCEC}\\
\midrule
 -&-& 0.594 $\pm$ 0.021  & 0.618 $\pm$ 0.041 &0.784 $\pm$ 0.021&0.643 $\pm$ 0.031 & 0.708 $\pm$ 0.065  \\
-&$\checkmark$ & 0.670 $\pm$ 0.028  & 0.629 $\pm$ 0.063 &0.806 $\pm$ 0.023&0.647 $\pm$ 0.042 & 0.713 $\pm$ 0.038 \\
$\checkmark$&$\checkmark$ &\textbf{ 0.671 $\pm$ 0.027  }& \textbf{0.658 $\pm$ 0.081} &\textbf{0.812 $\pm$ 0.028}&\textbf{0.655 $\pm$ 0.050} & \textbf{0.721 $\pm$ 0.057} \\
\bottomrule
\end{tabular}
}
\vspace{0em}
\label{table6}
\end{table*}

\begin{table*}[tp]
\centering
\vspace{-1.5em}
\caption{comparision of EM-based slide embedding with other methods in all datasets}
\resizebox{0.93\textwidth}{!}{%
\begin{tabular}{l|ccccc}
\toprule
\textbf{slide embedding.} &  \textbf{BLCA} & \textbf{BRCA} & \textbf{GBMLGG}& \textbf{LUAD}& \textbf{UCEC}\\
\midrule
-  & 0.670 $\pm$ 0.028  & 0.629 $\pm$ 0.063 &0.806 $\pm$ 0.023&0.647 $\pm$ 0.042 & 0.713 $\pm$ 0.038  \\
Rand  &0.657 $\pm$ 0.032&0.615 $\pm$ 0.043& 0.809 $\pm$ 0.027&0.651 $\pm$ 0.050 & 0.707 $\pm$ 0.051  \\
Cluster-Based  & 0.670 $\pm$ 0.038  & 0.636 $\pm$ 0.034 &0.812 $\pm$ 0.029 &0.644 $\pm$ 0.032&0.703 $\pm$ 0.038 \\
EM-Based(Ours)  & \textbf{0.671 $\pm$ 0.027}  & \textbf{0.658 $\pm$ 0.081} &\textbf{0.812 $\pm$ 0.028}&\textbf{0.655 $\pm$ 0.050} & \textbf{0.721 $\pm$ 0.057}  \\
\bottomrule
\end{tabular}
}
\vspace{-1.5em}
\label{table7}
\end{table*}

\vspace{1pt}\noindent\textbf{Weak Supervision with Limited Batch Sizes.} A second approach to survival prediction using deep learning is to consider discrete time intervals and model each interval using an independent output neuron. This formulation overcomes the need for large mini-batches and allows the model to be optimized using single observations during training. Specifically, given right-censored survival outcome data, we build a discrete-time survival model by partitioning the continuous time scale into non-overlapping bins: $[t_0,t_1), [t_1,t_2), [t_2,t_3), [t_3,t_4)$ based on the quartiles of survival time values of uncensored patients in each TCGA cohort. The discrete event time of each patient, indexed by $j$, with continuous event time $T_{j,cont}$ is then defined by:
\begin{align}
T_j & = r \ \mathrm{if} \ T_{j,cont} \in [t_r,t_{r+1}) \ \mathrm{for} \ r\in \left \{ 0,1,2,3 \right \}.
\tag{12}
\end{align}
Given the discrete-time ground truth label of the $j_{th}$ patient as $Y_j$. For a given patient with bag-level feature $h^L_j$ , the last layer of the network uses the sigmoid activation and models the hazard function defined as:
\begin{align}
\small
h(r|h_j^L)  = P(T_j  = r|T_j \ge r, h^L_j),
\tag{13}
\end{align}
which relates to the survival function through:
\begin{equation}
\begin{aligned}
\small
S(r|h^L_j) &= P(T_j > r|h^L_j)\\
                  &=\prod_{u=1}^{r}(1-h(u|h^L_j)).
\end{aligned}
\tag{14}
\end{equation}

\vspace{1pt}\noindent\textbf{Negative Likelihood Loss.} 
During training, we update the model parameters using the Negative Likelihood Loss (NLL) $\mathcal{L}_{NLL}$ for a discrete survival model, taking into account each patient’s binary censorship status ($c_j = 1$ if the patient lived past the end of the follow-up period and $c_j=0$ for patients who passed away during the recorded event time $T_j$):

\begin{equation}
\begin{aligned}
\small
\mathcal{L}_{NLL} = -l & = -c_j\cdot (S(Y_j|h_j^L)) \nonumber\\
       &- (1-c_j) \cdot log(S(Y_j-1|h^L_j)) \\
       &-(1-c_j) \cdot log(S(Y_j|h^L_j)). \nonumber
\end{aligned}
\tag{15}
\end{equation}

\begin{figure}
\centerline{\resizebox{3in}{2in}{\includegraphics{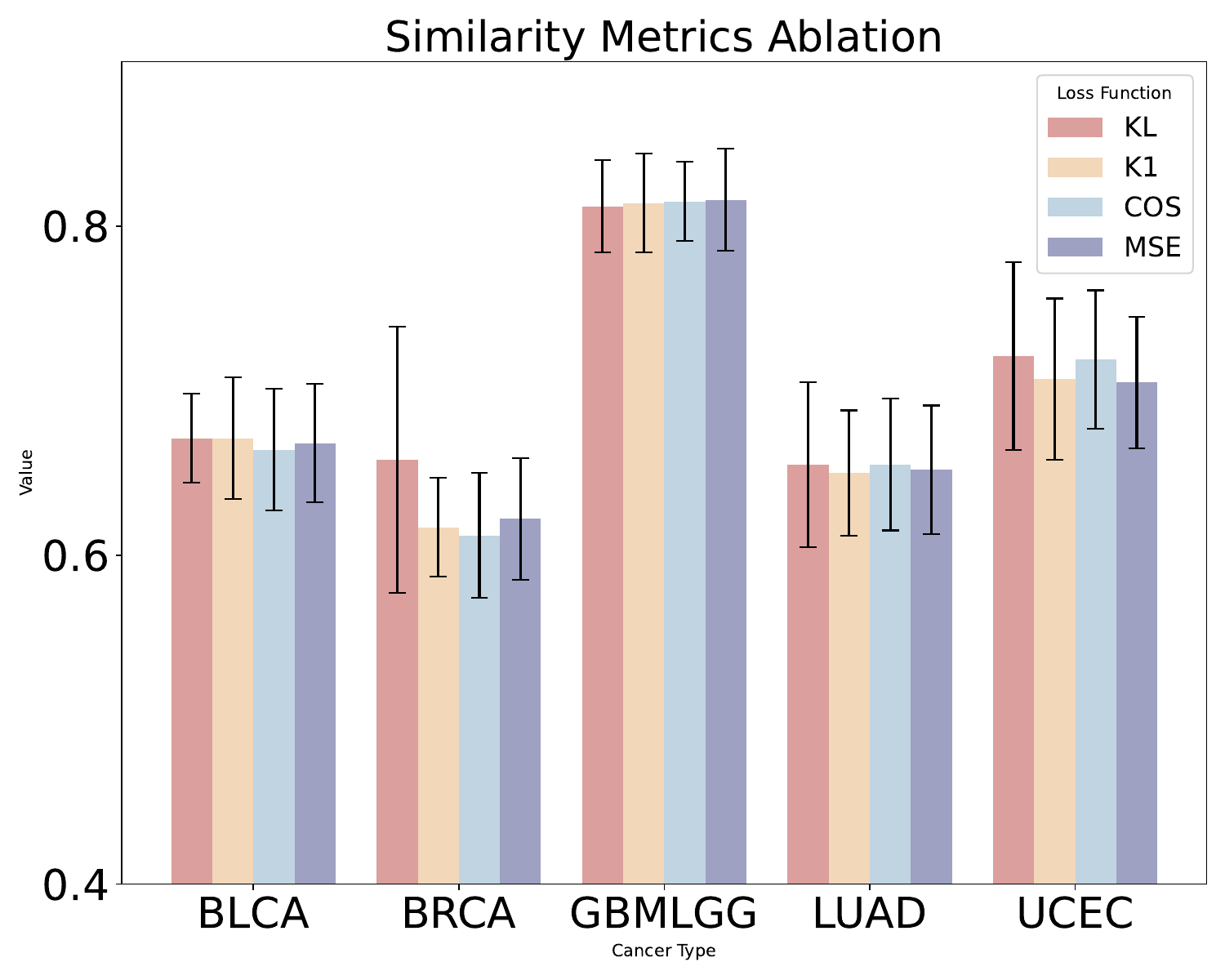}}}
\caption{Performance under different similarity metrics.}
\label{fig4}
\end{figure}

\subsection{Detailed Results of Ablation Studies}
\vspace{1pt}\noindent\textbf{Impacts of Similarity Metrics.}
In most works involving missing modality imputation or cross-modal learning, the KL divergence is commonly used as a similarity measure, and we continue to adopt this choice. In this part, we conduct some experiments to evaluate the performances under other common similarity metrics, i.e., MSE (mean squared error) loss, Minimum Absolute Error (L1 Norm), and Cosine similarity. The experimental results are shown in Fig. \ref{fig4}. KL divergence achieves the best performance in four of the five datasets, especially on brca, where it improves by $5.7\%$ compared to the second highest similarity measurement strategy of C-index.

\begin{figure*}[tp]
    \centering
    \includegraphics[width=0.85\linewidth]{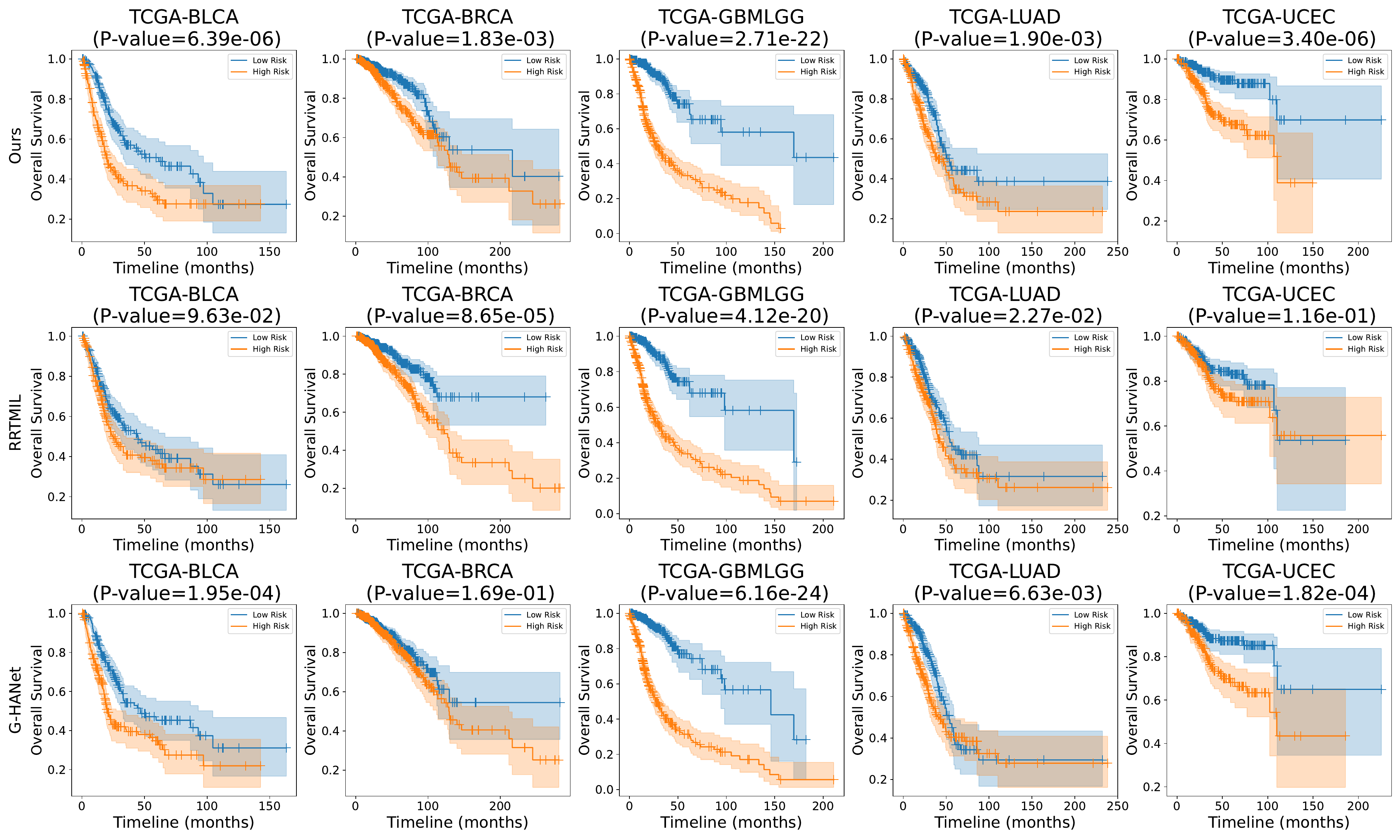}
    \vspace{-1em}
    \caption{Compare Kaplan-Meier analysis and Logrank test (p-value) with other methods.}
    \label{fig5}
    \vspace{-1.5em}
\end{figure*}

\vspace{1pt}\noindent\textbf{Multi-Branch Ablation Studies.}
In the main text, we have introduced the ablation studies of the WSI Learning Branch and the Genomics Guide Branch on three datasets (BLCA, BRCA, LUAD). Here, we expand the dataset results to fully demonstrate the detailed performance on all five datasets, which are comprehensively presented in Table \ref{table5}, \ref{table6}, and \ref{table7}. Similar to the main text, the best performance on each dataset is highlighted in the tables. The results show that the removal or replacement of any of our modules in the WSI Learning Branch and the Genomics Guide Branch leads to a degradation in model performance, thereby demonstrating the effectiveness and reliability of our method. In addition, in Table \ref{table8}, we also conducted an additional exploration of the performance of the attention module, which demonstrates the powerful performance of the cross-attention mechanism.

\vspace{1pt}\noindent\textbf{Multi-Ethnic Generalization Verification.} To verify the generalization ability of VGAT, We used the European and American race in TCGA-BRCA/GBMLGG as the training set and the remaining races as the test set to verify the generalization ability of the model. Then we compared it with SOTA models of various modalities. The results are shown in Table \ref{table9}. The performance values of all methods declined due to the extreme differences between the training set and the test set. Nevertheless, our method still remains the SOTA among WSI-based methods.

\subsection{Comparison of Patient Stratification Results.}
We compared the Kaplan-Meier analysis results of our VGAT model with the two best-performing methods in the WSI-based inference models, G-HANet and RRT-MIL. The results in Fig. \ref{fig5} clearly demonstrate that our p-values and curve visualizations are superior across the BLCA, LUAD, and UCEC datasets. Notably, in the BLCA dataset, our method achieved a p-value that is only $3.3\%$ of the second-best method, highlighting a significant advantage. This trend is consistent with the C-index results. In the UCEC dataset, where gene reconstruction methods outperformed all other approaches, achieving the best performance was particularly challenging. Despite this, our method still managed to perform the best, which underscores its strength. In the other two datasets, our results were consistently among the top, with only a marginal difference from the best performance. This indicates that VGAT has no significant flaws and performs well across all datasets.

\begin{table}[tp]
\centering
\vspace{-0.5em}
\caption{Coattn block ablation experiment}
\resizebox{0.47\textwidth}{!}{%
\begin{tabular}{l|ccccc}
\toprule
\textbf{Methods} &  \textbf{BLCA} &  \textbf{BRCA} &  \textbf{GBMLGG}&  \textbf{LUAD}&  \textbf{UCEC} \\
\midrule
VGAT(without Coatt)  & 0.645$\pm$ 0.089& 0.622$\pm$ 0.107& 0.797$\pm$ 0.058 & 0.639$\pm$ 0.072 &0.718$\pm$ 0.093  \\
VGAT(ours)  & \textbf{0.671$\pm$ 0.027}& \textbf{0.658$\pm$ 0.081}& \textbf{0.812$\pm$ 0.028} & \textbf{0.655$\pm$ 0.050} &\textbf{0.721$\pm$ 0.057}  \\

\bottomrule
\end{tabular}
}
\vspace{-1em}
\label{table8}
\end{table}

\begin{table}[tp]
\centering
\vspace{-0.5em}
\caption{Multi-ethnic generalization verification results}
\resizebox{0.47\textwidth}{!}{%
\begin{tabular}{l|ccccc}
\toprule
\textbf{Dataset} &  \textbf{SNNTrans} &\textbf{RRTMIL} &  \textbf{MOTCat} &  \textbf{G-HANet}&  \textbf{VGAT(ours)} \\
\midrule
BRCA  &0.608 & 0.573& 0.617& 0.577 & 0.605  \\
GBMLGG  &0.828 & 0.763& 0.807& 0.784 & 0.799  \\
\bottomrule 
\end{tabular}
}
\vspace{-1em}
\label{table9}
\end{table}

\subsection{Visualization of Vision Genomic Answering.}
To demonstrate interpretability, we visualize the attention values of all instances used in each WSI in the Pathology Transformer of VGAT, serving as a reference for its visual attention scope. Due to the late fusion of reconstructed gene embeddings, the regions of visual attention in the images will correlate with regions of gene expression enrichment, thereby exhibiting superior local attention capabilities compared to traditional WSI-based methods. As mentioned in the main text, our method degrades to TransMIL in the absence of VGA and ESE modules. Therefore, we also compare the visualization results of the same images using TransMIL to investigate the effectiveness of gene reconstruction.

\begin{figure*}[tp]
    \centering
    \includegraphics[width=\linewidth]{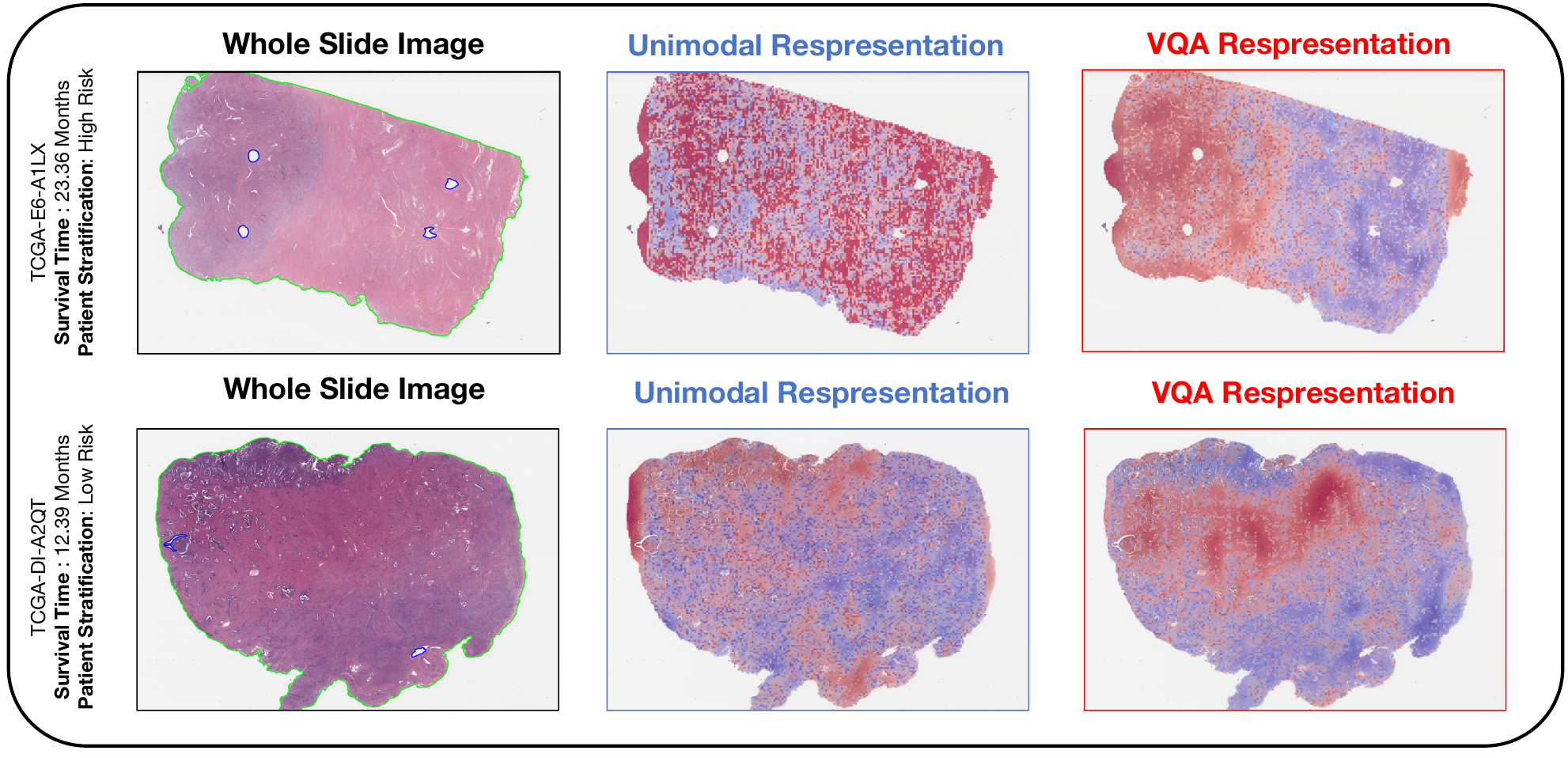}
    \vspace{-2em}
    \caption{Visualization Comparison Result 1}
    \label{fig6}
    \vspace{-1.5em}
\end{figure*}

\begin{figure*}[tp]
    \centering
    \includegraphics[width=\linewidth]{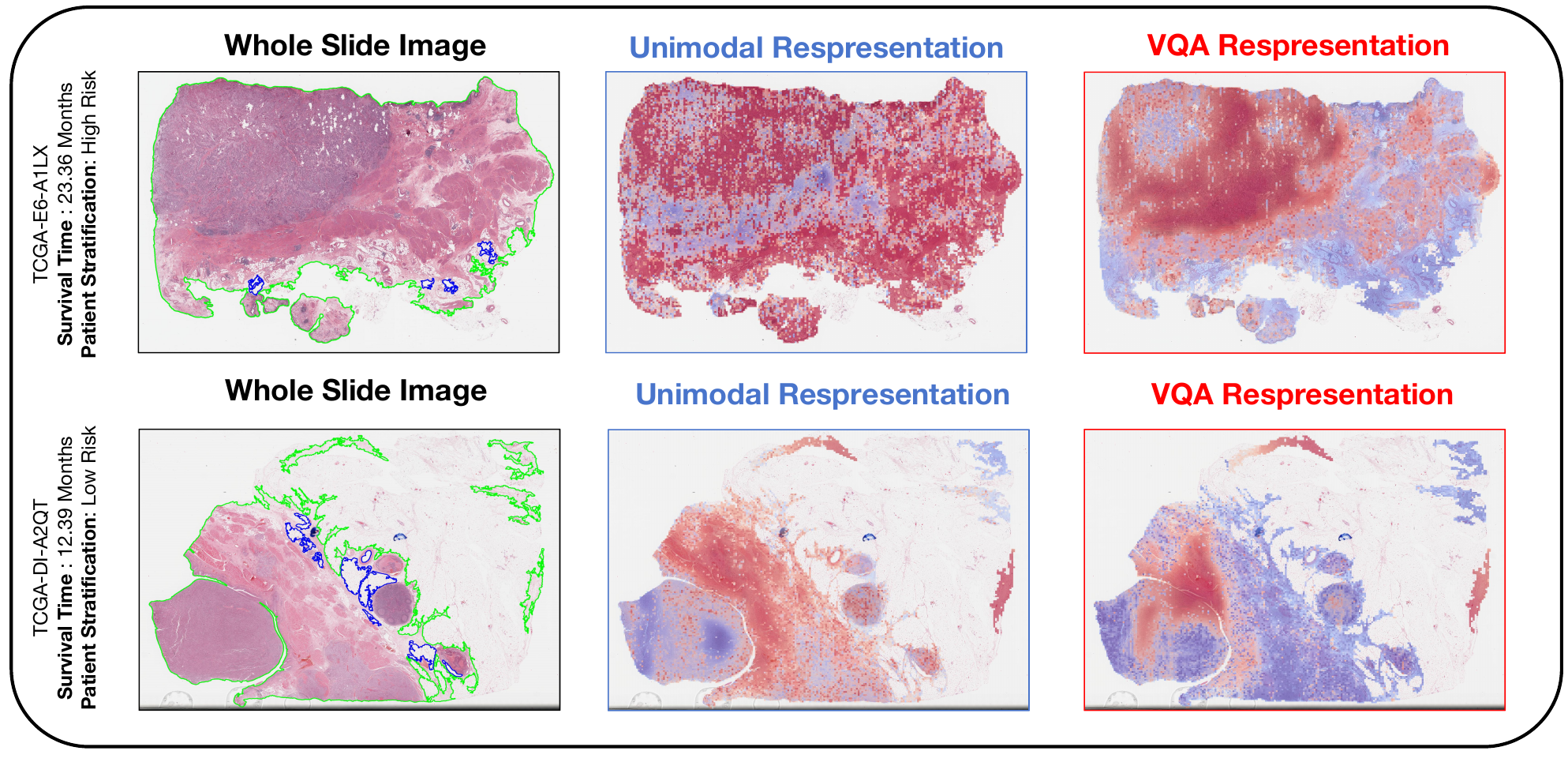}
    \vspace{-2em}
    \caption{Visualization Comparison Result 2}
    \label{fig7}
    \vspace{-1.5em}
\end{figure*}

In Fig. \ref{fig6} and \ref{fig7}, two sets of visualization results are presented as references. Each set includes one sample from high-risk patients and one sample from low-risk patients, along with their corresponding WSI, all sampled from the same dataset. The visualization results of the Pathology Transformer in the VGAT are presented as the performance under the VGA method, while the visualization results from TransMIL are presented as the traditional unimodal performance. These are respectively labeled with red and blue fonts and boxes.

Compared to traditional unimodal methods, the VGA method enables the visual module's attention to focus on smaller regions, effectively addressing redundant information in WSIs and concentrating on gene - enriched or tumor - microenvironment areas. Moreover, it shows that the reconstructed genes offer accurate guidance, validating the reliability of gene reconstruction and the entire VGAT framework. 

\end{document}